\documentstyle[epsf,aaai]{article}

\newcounter{diacount}
\newcounter{diacontinue}
\newcommand{\speaker}{A}
\newcommand{\listener}{B}
\newlength{\dialin}

\newcommand{\edialog}
   {\end{list}}

\newcommand{\edialogat}
   {\end{list}}

\newcommand{\bdialog}[2]
  {\renewcommand{\speaker}{#1}
   \renewcommand{\listener}{#2}
   \settowidth{\dialin}{\speaker}
   \begin{list}
         {}
         {\usecounter{diacount}
          \setlength{\labelwidth}{5em}
          \setlength{\rightmargin}{\leftmargin}
	  \addtolength{\leftmargin}{2em}}}

\newcommand{\bdialogcont}[2]
  {\renewcommand{\speaker}{#1}
   \renewcommand{\listener}{#2}
   \settowidth{\dialin}{\speaker}
   \setcounter{diacontinue}{\thediacount}
   \begin{list}
         {}
         {\usecounter{diacount}
	  \setcounter{diacount}{\thediacontinue}
          \setlength{\labelwidth}{5em}
          \setlength{\rightmargin}{\leftmargin}
	  \addtolength{\leftmargin}{2em}}}

\newcommand{\speakerlab}
  {
   \refstepcounter{diacount}
   \item[\hspace*{3em}(\arabic{diacount}) \speaker]}

\newcommand{\dialine}
  {
   \refstepcounter{diacount}
   \item[\hspace*{3em}(\arabic{diacount}) \hspace*{\dialin}]}

\makeatletter

\newcounter{subfigure}[figure]

\def\thesubfigure{(\alph{subfigure})\space}

\def\subcapsize{\footnotesize}

\def\subfigtopskip{10pt}

\def\subfigcapskip{10pt}

\def\subfigcapmargin{10pt}

\def\subfigure{%
  \leavevmode
  \@ifnextchar [%
    \@subfigure
    {\@subfigure[\@empty]}}

\long\def\@subfigure[#1]#2{%
  \stepcounter{subfigure}%
  \setbox\@tempboxa \hbox{#2}%
  \@tempdima=\wd\@tempboxa
  \vtop{%
    \vbox{
      \vskip\subfigtopskip
      \box\@tempboxa}
    \vskip\subfigcapskip
    \begingroup
      \@parboxrestore
      \setbox\@tempboxa
      \ifx #1\@empty
        \hbox{\subcapsize\strut\hfil}%
      \else
        \hbox{\subcapsize\strut\thesubfigure#1}%
      \fi
      \@tempdimb=-\subfigcapmargin
      \multiply\@tempdimb\tw@
      \advance\@tempdimb\@tempdima
      \hbox to\@tempdima{%
        \hfil
        \ifdim \wd\@tempboxa >\@tempdimb 
          \parbox{\@tempdimb}{\subcapsize\thesubfigure#1}%
        \else
          \box\@tempboxa
        \fi
        \hfil}
    \endgroup
  \vskip\subfigtopskip}}

\makeatother

\title{A Plan-Based Model for Response Generation in\\
       Collaborative Task-Oriented Dialogues\thanks{This material is
based upon work supported by the National Science Foundation under
Grant No. IRI-9122026.}}

\author{Jennifer Chu-Carroll \\
        Department of Computer Science\\
        University of Delaware\\
        Newark, DE 19716, USA\\
        E-mail: jchu@cis.udel.edu
        \And
        Sandra Carberry\\
        Department of Computer Science\\
        University of Delaware\\
        Newark, DE 19716, USA\\
        Visitor: Institute for Research in Cognitive Science\\
        University of Pennsylvania\\
        E-mail: carberry@cis.udel.edu}
\date{}

\begin{document}
\maketitle
\thispagestyle{empty}
\pagestyle{empty}

\begin{abstract}

This paper presents a plan-based architecture for response generation
in collaborative consultation dialogues, with emphasis on cases in
which the system (consultant) and user (executing agent) disagree.
Our work contributes to an overall system for collaborative
problem-solving by providing a plan-based framework that captures the
{\em Propose-Evaluate-Modify} cycle of collaboration, and by allowing
the system to initiate subdialogues to negotiate proposed additions to
the shared plan and to provide support for its claims. In addition,
our system handles in a unified manner the negotiation of proposed
domain actions, proposed problem-solving actions, and beliefs proposed
by discourse actions. Furthermore, it captures cooperative responses
within the collaborative framework and accounts for why questions are
sometimes never answered.
\end{abstract}

\section{Introduction}
\label{intro}

In collaborative expert-consultation dialogues, two participants
(executing agent and consultant) work together to construct a plan for
achieving the executing agent's domain goal. The executing agent and
the consultant bring to the plan construction task different knowledge
about the domain and the desirable characteristics of the resulting
domain plan.  For example, the consultant presumably has more
extensive and accurate domain knowledge than does the executing agent,
but the executing agent has knowledge about his particular
circumstances, intentions, and preferences that are either
restrictions on or potential influencers \cite{bra_ic90} of the domain
plan being constructed. In agreeing to collaborate on constructing the
domain plan, the consultant assumes a stake in the quality of the
resultant plan and in how the agents go about constructing it. For
example, a consultant in a collaborative interaction must help the
executing agent find the best strategy for constructing the domain
plan, may initiate additions to the domain plan, and must negotiate
with the executing agent when the latter's suggestions are not
accepted (rather than merely agreeing to what the executing agent
wants to do). Thus a collaborator is more than a cooperative
respondent.

In this paper, we present a plan-based architecture for response
generation in collaborative consultation dialogues, with emphasis on
cases in which the system and the user disagree. The model treats
utterances as proposals open for negotiation and only incorporates a
proposal into the shared plan under construction if both agents
believe the proposal to be appropriate. If the system does not accept
a user proposal, the system attempts to modify it, and natural
language utterances are generated as a part of this process. Since the
system's utterances are also treated as proposals, a recursive
negotiation process can ensue. This response generation architecture
has been implemented in a prototype system for a university advisement
domain.

\section{Modeling Collaboration}

In a collaborative planning process, conflicts in agents' beliefs must
be resolved as soon as they arise in order to prevent the agents from
constructing different plans. Hence, once a set of actions is proposed
by an agent, the other agent must first evaluate the proposal based on
his own private beliefs \cite{all_snlw91} and determine whether or not
to accept the proposal. If an agent detects any conflict which leads
him to reject the proposal, he should attempt to modify the proposal
to a form that will be accepted by both agents --- to do otherwise
is to fail in his responsibilities as a participant in collaborative
problem-solving. Thus, we capture collaboration in a {\em
Propose-Evaluate-Modify} cycle. This theory views the collaborative
planning process as a sequence of proposals, evaluations, and
modifications, which may result in a fully constructed shared plan
agreed upon by both agents. Notice that this model is essentially a
recursive one: the {\em Modify} action in itself contains a full
collaborative process --- an agent's proposal of a modification,
the other agent's evaluation of the proposal, and potential
modification to the modification!

We capture this theory in a plan-based system for response generation
in collaborative task-oriented interactions. We assume that the
current status of the interaction is represented by a tripartite
dialogue model \cite{lam_car_acl91} that captures intentions on three
levels: domain, problem-solving, and discourse.  The domain level
contains the domain plan being constructed for later execution. The
problem-solving level contains the agents' intentions about how to
construct the domain plan, and the discourse level contains the
communicative plan initiated to further their joint problem-solving
intentions.

Each utterance by a participant constitutes a {\em proposal} intended
to affect the shared model of domain, problem-solving, and discourse
intentions. For example, relating a user's query such as {\em Who is
teaching AI?} to an existing tripartite model might require inferring
a chain of domain actions that are not already part of the plan,
including {\em Take-Course(User,AI)}. These inferred actions explain
{\em why} the user asked the question and are actions that the user is
implicitly proposing be added to the plan. In order to capture the
notion of {\em proposals} vs. {\em shared plans} in a collaborative
planning process, we separate the dialogue model into an {\em existing
model}, which consists of a shared plan agreed upon by both agents,
and the {\em proposed additions}, which contain newly inferred
actions.  Furthermore, we augment Lambert's plan recognition algorithm
\cite{lam_car_acl92} with a simplified version of Eller's relaxation
algorithm \cite{ell_car_umuai92} to recognize ill-formed plans.

We adopt a plan-based mechanism because it is general and easily
extendable, allows the same declarative knowledge about collaborative
problem-solving to be used both in generation and understanding, and
allows the recursive nature of our theory to be represented by
recursive meta-plans.  This paper focuses on one component of our
model, the {\bf arbitrator}, which performs the {\em Evaluate} and
{\em Modify} actions in the {\em Propose-Evaluate-Modify} cycle of
collaboration.

\section{The Arbitration Process}

A {\em proposal} consists of a chain of actions for addition to the
shared plan.  The {\bf arbitrator} evaluates a proposal and determines
whether or not to accept it, and if not, modifies the original
proposal to a form that will potentially be accepted by both
agents. The {\bf arbitrator} has two subcomponents, the {\bf
evaluator} and the {\bf modifier}, and has access to a library of
generic recipes for performing actions\footnote{A recipe
\cite{pol_acl86} is a template for performing an action. It encodes
the {\em preconditions} for an action, the {\em effects} of an action,
the {\em subactions} comprising the body of an action, etc.}.

\subsection{The Evaluator}

\looseness=-1000
A collaborative agent, when presented a proposal, needs to decide
whether or not he believes that the proposal will result in a valid
plan and will produce a reasonably efficient way to achieve the
high-level goal. Thus, the {\bf evaluator} should check for two types
of discrepancies in beliefs: one that causes the proposal to be viewed
by the system as invalid \cite{pol_acl86}, and one in which the system
believes that a better alternative to the user's proposal exists
\cite{josetal_aaai84,vb_acl87}. Based on this evaluation,
the system determines whether it should accept the user's proposal,
causing the proposed actions to be incorporated into the existing
model, or should reject the proposal, in which case a negotiation
subdialogue will be initiated.

The processes for detecting conflicts and better alternatives start at
the top-level proposed action, and are interleaved because we intend
for the system to address the highest-level action disagreed upon by
the agents.  This is because it is meaningless to suggest, for
example, a better alternative to an action when one believes that its
parent action is infeasible.

\subsubsection{Detecting Conflicts About Plan Validity}

Pollack argues that a plan can fail because of an {\em infeasible
action} or because the plan itself is {\em ill-formed}
\cite{pol_acl86}. An action is {\em infeasible} if it cannot be
performed by its agent; thus, the {\bf evaluator} performs a {\em
feasibility} check by examining whether the applicability conditions
of the action are satisfied and if its preconditions can be
satisfied\footnote{Applicability conditions are conditions that must
already be satisfied in order for an action to be reasonable to
pursue, whereas an agent can try to achieve unsatisfied preconditions.
Our evaluator considers a precondition satisfiable if there exists an
action which achieves the precondition and whose applicability
conditions are satisfied. Thus only a cursory evaluation of
feasibility is pursued at this stage of the planning process, with
further details considered as the plan is worked out in depth. This
appears to reflect human interaction in naturally occuring
dialogues.}. A plan is considered {\em ill-formed} if child actions do
not contribute to their parent action as intended; hence, the
evaluator performs a {\em well-formedness} check to examine, for each
pair of parent-child actions in the proposal, whether the {\em
contributes} relationship holds between them\footnote{Much of the
information needed for the feasibility and well-formedness checks will
be provided by the plan-recognition system that identified the actions
comprising the proposal.}. The well-formedness check is performed
before the feasibility check since it is reasonable to check the
relationship between an action and its parent before examining the
action itself.

\subsubsection{Detecting Sub-Optimal Solutions}

It is not sufficient for the system, as a collaborator, to accept or
reject a proposal merely based on its validity. If the system knows of
a substantially superior alternative to the proposal, but does not
suggest it to the user, it cannot be said to have fulfilled its
responsibility as a collaborative agent; hence the system must model
user characteristics in order to best tailor its identification of
sub-optimal plans to individual users. Our system maintains a user
model that includes the user's {\em preferences}. A preference
indicates, for a particular user, the preferred value of an attribute
associated with an object and the strength of this preference. The
preferences are represented in the form, prefers(\_user,
\_attribute(\_object, \_value), \_action, \_strength), which indicates
that \_user has a \_strength preference that the attribute \_attribute
of \_object be \_value when performing \_action. For instance, {\em
prefers(UserA, Difficulty(\_course, easy), Take-Course, weak)}
indicates that UserA has a weak preference for taking easy courses. A
companion paper describes our mechanism for recognizing user
preferences during the course of a dialogue \cite{elzetal_um94}.

Suppose that the {\bf evaluator} must determine whether an action
$A_i$ (in a chain of proposed actions $A_1,\ldots,A_i,\ldots,A_n$) is
the best way of performing its parent action $A_{i+1}$. We will limit
our discussion to the situation in which there is only one generic
action (such as {\em Take-Course}) that achieves $A_{i+1}$, but there
are several possible instantiations of the parameters of the action
(such as {\em Take-Course(UserA,CS601)} and {\em
Take-Course(UserA,CS621)}).

\paragraph{The Ranking Advisor}

The ranking advisor's task is to determine how best the parameters of
an action can be instantiated, based on the user's preferences.  For
each object that can instantiate a parameter of an action (such as
CS621 instantiating \_course in {\em Take-Course(UserA,\_course)}),
the {\bf evaluator} provides the ranking advisor with the values of
its attributes (e.g., {\em Difficulty(CS621,difficult)}) and the
user's preferences for the values of these attributes (e.g., {\em
prefers(UserA, Difficulty(\_course,moderate), Take-Course, weak)}).

Two factors should be considered when ranking the candidate
instantiations: the {\em strength of the preference} and the {\em
closeness of the match}. The strength of a preference\footnote{We
model six degrees each of positive and negative preferences based on
the conversational circumstances and the semantic representation of
the utterance used to express the preferences \cite{elzetal_um94}.}
indicates the {\em weight} that should be assigned to the
preference. The closeness of the match ({\em exact, strong, weak,} or
{\em none}) measures how well the actual and the preferred values of
an attribute match. It is measured based on the {\em distance} between
the two values where the unit of measurement differs depending on the
type of the attribute. For example, for attributes with discrete
values ({\em difficulty} of a course can be {\em very-difficult,
difficult, moderate, easy}, or {\em very-easy}), the match between
{\em difficult} and {\em moderate} will be {\em strong}, while that
between {\em difficult} and {\em easy} will be {\em weak}. The
closeness of the match must be modeled in order to capture the fact
that if the user prefers difficult courses, a moderate course will be
considered preferable to an easy one, even though neither of them
exactly satisfies the user's preference.

For each candidate instantiation, the ranking advisor assigns
numerical values to the strength of the preferences for the relevant
attributes and computes the closeness of each match. A weight is
computed for each candidate instantiation by summing the products of
corresponding terms of the strength of a preference and the closeness
of a match. The instantiation with the highest weight is considered
the {\em best} instantiation for the action under consideration. Thus,
the selection strategy employed by our ranking advisor corresponds to
an {\em additive model} of human decision-making \cite{ree_cog82}.

\paragraph{Example}

We demonstrate the ranking advisor by showing how two different
instantiations, CS601 and CS621, of the {\em Take-Course} action are
ranked.  Figure~\ref{info} shows the relevant domain knowledge and
user model information.

\begin{figure}
\footnotesize
\begin{tabbing}
111 \= Prefers( \= \kill
{\bf Domain Knowledge:}\\
\> Teaches(Smith,CS601)\\
\> Meets-At(CS601,2-3:15pm)\\
\> Difficulty(CS601,difficult)\\
\> Workload(CS601,moderate)\\
\> Offered(CS601)\\
\> Content(CS601,\{formal-languages, grammar\})\\ \\
\> Teaches(Brown,CS621)\\
\> Meets-At(CS621,8-9:15am)\\
\> Difficulty(CS621,difficult)\\
\> Workload(CS621,heavy)\\
\> Offered(CS621)\\
\> Content(CS621,\{algorithm-design, complexity-theory\})\\ \\
{\bf User Model Information:}\\
\> Prefers(UserA, Meets-At(\_course,10am-5pm), \\
\> \> \_action, very-strong)\\
\> Prefers(UserA, Difficulty(\_course,moderate), \\
\> \> Take-Course, weak)\\
\> Prefers(UserA, Workload(\_course,heavy), \\
\> \> Take-Course, low-moderate)\\
\> Prefers(UserA, Content(\_course,formal-languages), \\
\> \> Take-Course,strong)
\end{tabbing}
\caption{System's Knowledge and User Model Information}
\label{info}
\vspace{1ex}
\hrule
\end{figure}

The ranking advisor matches the user's preferences against the domain
knowledge for each of CS601 and CS621. The attributes that will be
taken into account are the ones for which the user has indicated
preferences. For each attribute, the advisor records the {\em strength
of the preference} and the {\em closeness of the match} for each
instantiation. For instance, in considering the attribute {\em
workload}, the strength of the preference will be {\em low-moderate},
and the closeness of the match will be {\em strong} and {\em exact}
for CS601 and CS621, respectively.  Table~\ref{compare} shows a
summary of the strength of the preferences and the closeness of the
matches for the relevant attributes for both instantiations.
Numerical values are then assigned and used to calculate a final
weight for each candidate. In this example, the normalized weight for
CS601 is $43/48$ and that for CS621 is $29/48$; therefore, CS601 is
considered a substantially better instantiation than CS621 for the
{\em Take-Course} action for UserA.

\begin{table}
\footnotesize
\begin{center}
\begin{tabular}{|l|lr|lr|r|}
\hline
{\bf CS601} & \multicolumn{2}{c|}{Preference-Strength} &
\multicolumn{2}{c|}{Match} & \\ \hline
Meets-At & very-strong & 6 & exact & 3 & 18 \\ \hline
Difficulty & weak & 2 & strong & 2 & 4 \\ \hline
Workload & low-moderate & 3 & strong & 2 & 6 \\ \hline
Content & strong & 5 & exact & 3 & 15 \\ \hline
& & & & & 43 \\ \hline
\end{tabular}
\end{center}
\begin{center}
\begin{tabular}{|l|lr|lr|r|}
\hline
{\bf CS621} & \multicolumn{2}{c|}{Preference-Strength} &
\multicolumn{2}{c|}{Match} & \\ \hline
Meets-At & very-strong & 6 & weak & 1 & 6\\ \hline
Difficulty & weak & 2 & strong & 2 & 4\\ \hline
Workload & low-moderate & 3 & exact & 3 & 9 \\ \hline
Content & strong & 5 & strong & 2 & 10 \\ \hline
& & & & & 29 \\ \hline
\end{tabular}
\end{center}
\caption{The Strengths of Preferences and Matches \label{compare}}
\hrule
\end{table}

\subsection{The Modifier}

\looseness=-1000
The {\bf modifier} is invoked when a proposal is rejected. Its task is
to modify the proposal to a form that will potentially be accepted by
both agents. The process is controlled by the {\em Modify-Proposal}
action, which has four specializations: 1) {\em Correct-Node}, for
when the proposal is infeasible, 2) {\em Correct-Relation}, for when
the proposal is ill-formed, 3) {\em Improve-Action}, for when a better
generic action is found, and 4) {\em Improve-Parameter}, for when a
better instantiation of a parameter is found. Each specialization
eventually decomposes into some primitive action which modifies the
proposal. However, an agent will be considered uncooperative if he
modifies a proposed shared plan without the collaborating agent's
consent; thus, the four specializations share a common precondition
--- that the discrepancies in beliefs must be {\em squared away}
\cite{jos_mk82} before any modification can take place. It is the
attempt to satisfy this precondition that causes the system to
generate natural language utterances to accomplish the change in the
user's beliefs.

Figure~\ref{recipes} shows two problem-solving recipes, {\em
Correct-Relation} and {\em Modify-Relation}, the latter being a
subaction of the former. The applicability conditions of {\em
Correct-Relation} indicate that it is applicable when the agents, \_s1
and \_s2, disagree on whether a particular relationship (such as {\em
contributes}) holds between two actions (\_node1 and \_node2) in the
proposal. The applicability condition and precondition of {\em
Modify-Relation} show that the action can only be performed if both
\_s1 and \_s2 believe that the relationship \_rel does not hold
between \_node1 and \_node2; in other words, the conflict between \_s1
and \_s2 must have been resolved. The attempt to satisfy this
precondition causes the system to invoke discourse actions to modify
the user's beliefs, which can be viewed as initiating a negotiation
subdialogue to resolve a conflict. If the user accepts the system's
beliefs, thus satisfying the precondition of {\em Modify-Relation},
the original dialogue model can be modified; however, if the user
rejects the system's beliefs, he will invoke the {\em Modify-Proposal}
action to revise the system's suggested modification of his original
proposal.

\begin{figure}
\footnotesize
\begin{tabbing}
Preconditions: \= Modify-Relation( \= \kill
Action: \> {\em Correct-Relation}(\_s1, \_s2, \_proposed)\\
Type: \> Decomposition\\
Appl Cond: \> believe(\_s1, $\lnot$holds(\_rel,\_node1,\_node2))\\
\> believe(\_s2, holds(\_rel,\_node1,\_node2))\\
Constraints: \> error-in-plan(\_relation, \_proposed)\\
\> name-of (\_relation, \_rel)\\
\> parent-node(\_relation, \_node2)\\
\> child-node(\_relation, \_node1)\\
Body: \> Modify-Relation(\_s1, \_s2, \_proposed,\\
\> \> \_rel, \_node1, \_node2)\\
\> Insert-Correction(\_s1, \_s2, \_proposed)\\
Effects: \> modified(\_proposed)\\
Goal: \> well-formed(\_proposed)\\ \\
Action: \> {\em Modify-Relation}(\_s1, \_s2, \_proposed,\\
\> \> \_rel, \_node1, \_node2)\\
Type: \> Specialization\\
Appl Cond: \> believe(\_s1, $\lnot$holds(\_rel,\_node1,\_node2))\\
Preconditions: \> believe(\_s2, $\lnot$holds(\_rel,\_node1,\_node2))\\
Body: \> Remove-Node(\_s1, \_s2, \_proposed, \_node1)\\
\> Alter-Node(\_s1, \_s2, \_proposed, \_node1)\\
Effects: \> modified(\_proposed)\\
Goal: \> modified(\_proposed)
\end{tabbing}
\caption{{\em Correct-Relation} and {\em Modify-Relation} Recipes}
\label{recipes}
\vspace{1ex}
\hrule
\end{figure}

In order to retain as much of the original proposal as possible when
modifying a proposal, {\em Modify-Relation} has two specializations:
{\em Remove-Node} and {\em Alter-Node}.  The former is selected if the
action itself is inappropriate, and will cause the action to be
removed from the dialogue model. The latter is chosen if a parameter
is inappropriately instantiated, in which case the action will remain
in the dialogue model and the problematic parameter will be left
uninstantiated.

\subsection{Example of Correcting an Invalid Proposal}

\begin{figure}
\centerline{\epsfysize=4.2in\epsffile{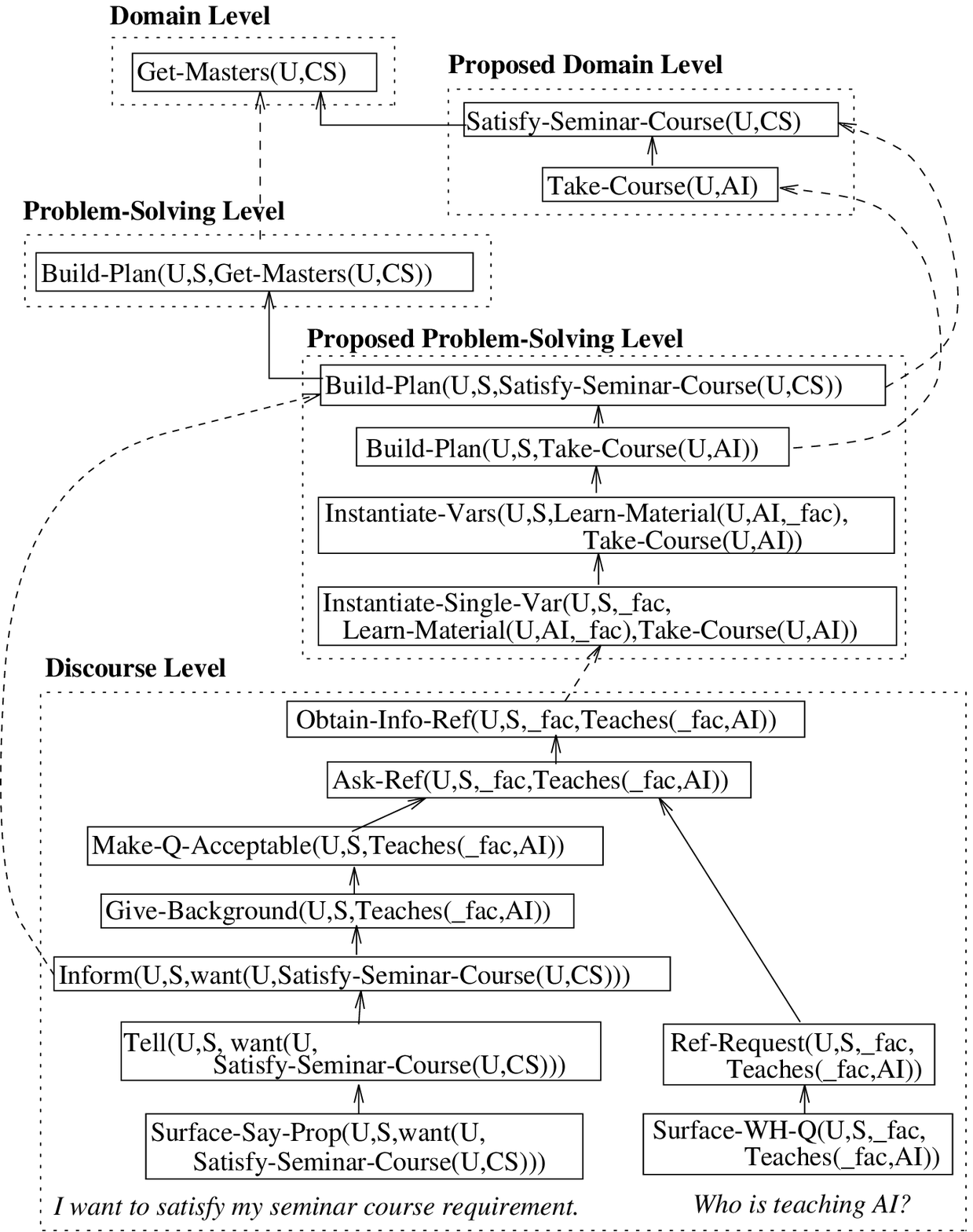}}
\caption{The Dialogue Model for Utterances
(\protect{\ref{seminar}})-(\protect{\ref{teach_ai}})}
\label{seminar_course}
\vspace{1ex}
\hrule
\end{figure}

Suppose earlier dialogue suggests that the user has the goal of
getting a Master's degree in CS ({\em Get-Masters(U,CS)}).
Figure~\ref{seminar_course} illustrates the dialogue model that would
result from the following utterances.

\bdialog{U:}{S:}
\em
\speakerlab \label{seminar} I want to satisfy my seminar course
requirement.
\dialine \label{teach_ai} Who is teaching AI?
\edialog

The evaluation process, which determines whether or not to accept the
proposal, starts at the top-level proposed domain action, {\em
Satisfy-Seminar-Course(U,CS)}. Suppose the system believes that {\em
Satisfy-Seminar-Course(U,CS)} contributes to {\em Get-Masters(U,CS)},
that U can perform {\em Satisfy-Seminar-Course(U,CS)}, and that there
is no better alternative to the instantiation of {\em
Satisfy-Seminar-Course}. The {\bf evaluator} then checks its child
action {\em Take-Course(U,AI)}. The system's recipe library indicates
that {\em Take-Course(U,AI)} does not contribute to {\em
Satisfy-Seminar-Course(U,CS)}, since it believes that AI is {\em not}
a seminar course, causing the proposal to be rejected.

\begin{figure}
\leavevmode
\centerline{\epsfysize=3.2in\epsffile{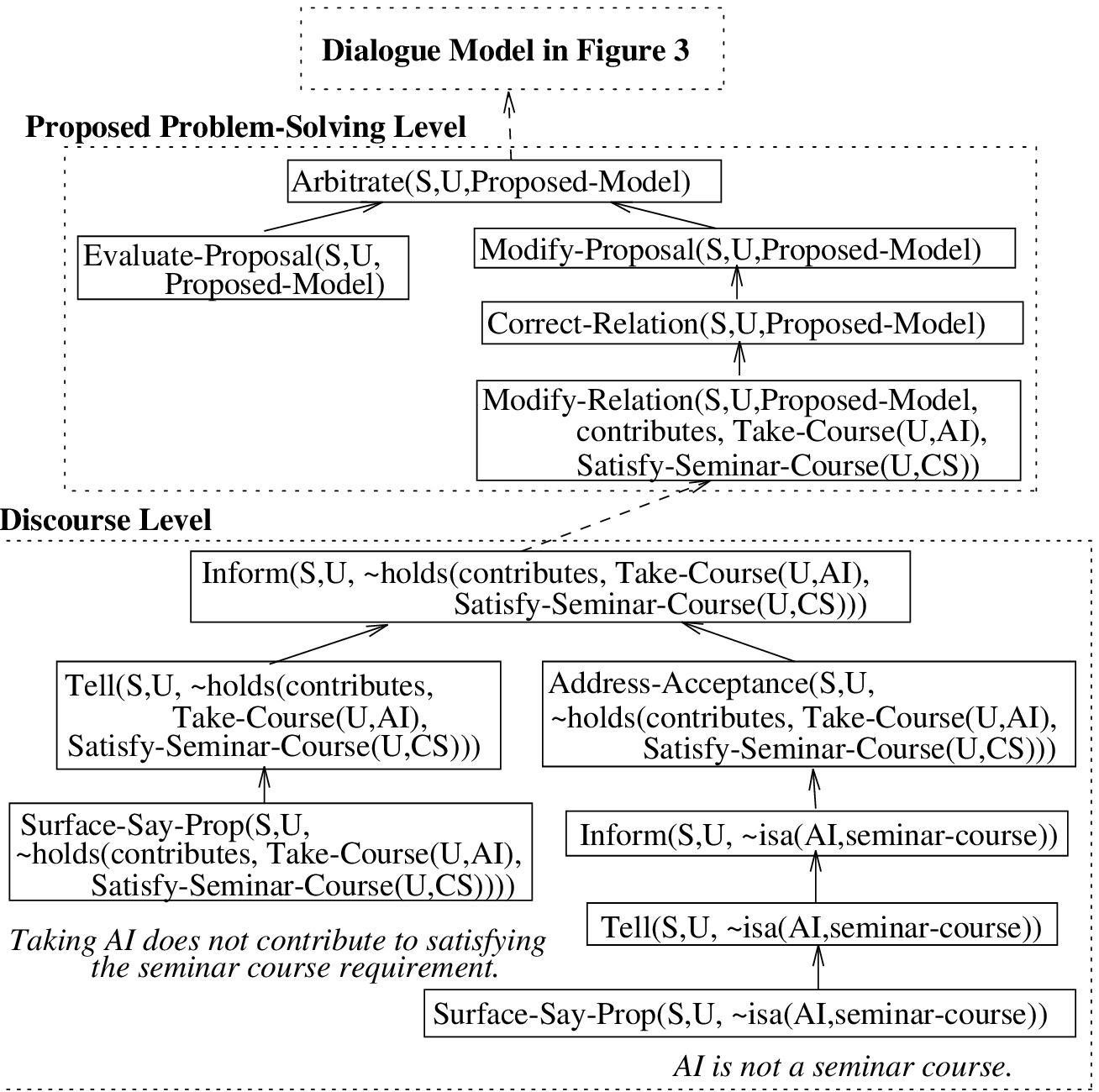}}
\caption{Responding to Implicitly-Conveyed Conflicts}
\label{correct_ai}
\vspace{1ex}
\hrule
\end{figure}

The {\bf modifier} performs the {\em Modify-Proposal} action, which
selects as its specialization {\em Correct-Relation}, because the
rejected proposal is ill-formed.  Figure~\ref{correct_ai} shows the
arbitration process and how {\em Correct-Relation} is expanded. Notice
that the arbitration process (the problem-solving level in
Figure~\ref{correct_ai}) operates on the entire dialogue model in
Figure~\ref{seminar_course}, and therefore is represented as
meta-level problem-solving actions.  In order to satisfy the
precondition of {\em Modify-Relation}, the system invokes the discourse
action {\em Inform} as an attempt to change the user's belief (in this
case, to achieve {\em believe(U,$\lnot$holds(contributes,
Take-Course(U,AI), Satisfy-Seminar-Course(U,CS)))}). The {\em Inform}
action further decomposes into two actions, one which tells the user
of the belief, and one which provides support for the claim.  This
process will generate the following two utterances:

\bdialogcont{S:}{}
\em
\speakerlab Taking AI does not contribute to satisfying the seminar
course requirement.
\dialine AI is not a seminar course.
\edialog

If the user accepts the system's utterances, thus satisfying the
precondition that the conflict be resolved, {\em Modify-Relation} can
be performed and changes made to the dialogue model. In this example,
the proposal is rejected due to an inappropriate instantiation of the
parameter \_course; thus {\em Modify-Relation} will select {\em
Alter-Node} as a specialization to replace all instances of AI in
the dialogue model with a variable. This variable can be
reinstantiated by {\em Insert-Correction}, the second subaction of
{\em Correct-Relation}.

Assuming that the system and the user encounter no further conflict in
reinstantiating the variable, the arbitration process at the
meta-level is completed and the original dialogue is returned to. The
proposed additions now consist of actions agreed upon by both agents
and will therefore be incorporated into the existing model. Notice
that our model separates the negotiation subdialogue (captured at the
meta level) from the original dialogue while allowing the same
plan-based mechanism to be used at both levels. It also accounts for
why the user's original question about the instructor of AI is never
answered --- a conflict was detected that made the question
superfluous.  Thus certain situations in which questions fail to be
answered can be accounted for by the collaborative process rather than
being viewed as a violation of cooperative behaviour.

\subsection{Example of Suggesting Better Alternatives}

Consider the following utterances, whose dialogue model has the same
structure as that for utterances (\ref{seminar}) and (\ref{teach_ai})
(Figure~\ref{seminar_course}).

\bdialogcont{U:}{}
\em
\speakerlab \label{theory} I want to satisfy my theory course
requirement.
\dialine \label{cs621} Who is teaching CS621?
\edialog

For space reasons, we skip ahead in the evaluation process to the
optimality check for {\em Take-Course(U,CS621)}.  There are two
instantiations of \_course that satisfy the constraints specified in
the recipe for {\em Satisfy-Theory-Course}: CS601 and CS621. These are
ranked by the ranking advisor based on the user's preferences,
summarized in Table~\ref{compare}, which suggests that CS601 is a
substantially better alternative to CS621. Thus, {\em
Improve-Parameter} is selected as a specialization of {\em
Modify-Proposal}. Similar to the previous example, the {\em Inform}
discourse action will be invoked as an attempt to resolve the
discrepancies in beliefs between the two agents, which would lead to
the generation of the following utterances:

\bdialogcont{S:}{}
\em
\speakerlab \label{cs601_better} CS601 is a better alternative than
CS621.
\dialine \label{cs601_why} CS601 meets at 2pm and involves formal
languages and grammar.
\edialog

\noindent Notice that utterance (\ref{cs601_why}) provides supporting
evidence for the claim in (\ref{cs601_better}), and is obtained by
comparing the sets of information used by the ranking advisor
(Table~\ref{compare}) and selecting the features that contribute most
to making CS601 preferable to CS621.

\section{The Belief Level}

We showed how our {\bf arbitrator} detects and resolves conflicts at
the domain level. Our goal, however, is to develop a mechanism that
can handle negotiations at the domain, problem-solving, and discourse
levels in a uniform fashion.  The process can be successfully applied
to the problem-solving level because both the domain and
problem-solving levels represent actions that the agents propose to do
(at a later point in time for the domain level and at the current time
for the problem-solving level); however, the discourse level actions
are actions that are {\em currently being executed}, instead of {\em
proposed for execution}. This causes problems for the modification
process, as illustrated by the following example.

\bdialogcont{U:}{S:}
\em
\speakerlab \label{ai} I want to take AI.
\dialine \label{brown} Dr. Brown is teaching AI,
\dialine \label{professor} since he is a full professor.
\edialog

\noindent Utterance (\ref{professor}) provides support for
(\ref{brown}), which supports (\ref{ai}). However, if the system
believes that whether one is a full professor has no relation to
whether or not he teaches AI, the system and the user have a conflict
as to whether (\ref{professor}) supports (\ref{brown}). Problems will
arise if the system convinces the user that Dr. Brown teaches AI
because that is his area of specialty, not because he is a full
professor, and attempts to modify the dialogue model by replacing the
{\em Inform} action that represents (\ref{professor}) with one that
conveys {\em specializes(Brown,AI)}.  This modification is
inappropriate because it indicates that the user informed the system
that Dr. Brown specializes in AI, which never happened in the first
place.  Therefore, we argue that instead of applying the arbitration
process to the discourse level, it should be applied to the beliefs
proposed by the discourse actions.

\begin{figure}
\leavevmode
\centerline{\epsfysize=2.8in\epsffile{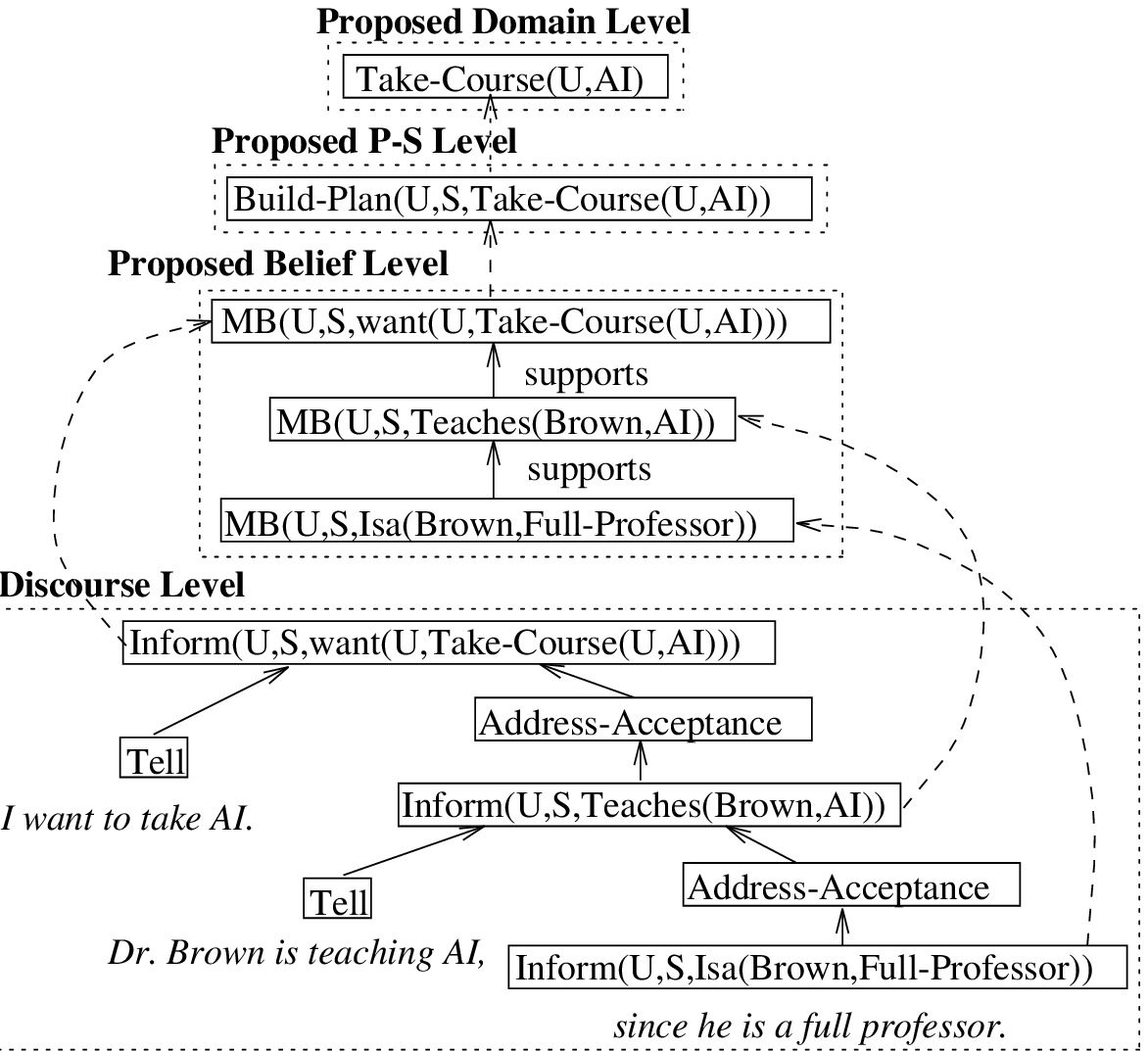}}
\caption{The Four-Level Model for Utterances
(\protect{\ref{ai}})-(\protect{\ref{professor}})}
\label{belief}
\vspace{1ex}
\hrule
\end{figure}

In order to preserve the representation of the discourse level, and to
handle the kind of conflict shown in the previous example, we expand
the dialogue model to include a {\em belief} level. The belief level
captures domain-related beliefs proposed by discourse actions as well
as the relationship amongst them. For instance, an {\em Inform} action
proposes a mutual belief (MB) of a proposition and an {\em
Obtain-Info-Ref} action proposes that both agents come to know the
referent (Mknowref) of a parameter. Thus, information captured at the
belief level consists not of actions, as in the other three levels,
but of beliefs that are to be achieved, and belief relationships, such
as {\em support}, {\em attack}, etc.

\paragraph{Discourse Level Example Revisited}

Figure~\ref{belief} outlines the dialogue model for utterances
(\ref{ai})-(\ref{professor}) with the additional belief level.  Note
that each {\em Inform} action at the discourse level proposes a mutual
belief, and that {\em supports} relationships (inferred from {\em
Address-Acceptance}) are proposed between the mutual beliefs.

The evaluation process starts at the proposed domain level. Suppose
that the system believes that both {\em Take-Course(U,AI)} and {\em
Build-Plan(U,S,Take-Course(U,AI))} can be performed. However, an
examination of the proposed belief level causes the proposal to be
rejected because the system does not believe that Dr. Brown being a
full professor supports the fact that he teaches AI. Thus, {\em
Correct-Relation} is selected as the specialization of {\em
Modify-Proposal} in order to resolve the conflict regarding this {\em
supports} relationship. Again in order to satisfy the precondition of
modifying the proposal, the system invokes the {\em Inform} action
which would generate the following utterance:

\bdialogcont{S:}{}
\em
\speakerlab Dr. Brown being a full professor does not provide support
for him teaching AI.
\edialog

Thus, with the addition of the belief level, the {\bf arbitrator} is
able to capture the process of evaluating and modifying proposals in a
uniform fashion at the domain, problem-solving, and belief levels. An
additional advantage of the belief level is that it captures the
beliefs conveyed by the discourse level, instead of {\em how} they are
conveyed (by an {\em Inform} action, by expressing doubt, etc.).

\section{Related Work}

Allen \shortcite{all_snlw91} proposed different plan modalities that
capture the shared and individual beliefs during collaboration, and
Grosz, Sidner and Lochbaum \cite{gro_sid_ic90,loc_acl91} proposed a
SharedPlan model for capturing intentions during a collaborative
process. However, they do not address response generation during
collaboration. Litman and Allen \shortcite{lit_all_cs87} used
discourse meta-plans to
handle correction subdialogues. However, their Correct-Plan only
addressed cases in which an agent adds a repair step to a pre-existing
plan that does not execute as expected. Thus their meta-plans do not
handle correction of proposed additions to the dialogue model, since
this generally does not involve adding a step to the proposal.
Furthermore, they were only concerned with understanding utterances, not
with generating appropriate responses. Heeman and Hirst
\shortcite{hee_hir_tr92} and
Edmonds \shortcite{edm_tr93} use meta-plans to account for
collaboration, but their mechanisms are limited to understanding and
generating referring expressions.  Although Heeman is extending his
model to account for collaboration in task-oriented dialogues
\cite{hee_aaaisym93}, his extension is limited to the recognition of
actions in such dialogues. Guinn and Biermann
\shortcite{gui_bie_ijcaiws93} developed a model of collaborative
problem-solving which attempts to resolve conflicts between agents
regarding the best path for achieving a goal. However, their work has
concentrated on situations in which the user is trying to execute a
task under the system's guidance rather than those where the system
and user are collaboratively developing a plan for the user to execute
at a later point in time.

Researchers have utilized plan-based mechanisms to generate natural
language responses, including explanations
\cite{moo_par_cl93,may_ijmms92,caw_ijmms93}. However, they only handle
cases in which the user fails to understand the system, instead of
cases in which the user {\em disagrees} with the system. Maybury
\shortcite{may_aaai93} developed plan operators for persuasive utterances,
but does not provide a framework for negotiation of conflicting views.

In suggesting better alternatives, our system differs from van Beek's
\shortcite{vb_acl87} in a number of ways. The most significant are
that our system dynamically recognizes user preferences
\cite{elzetal_um94}, takes into account both the strength of the
preferences and the closeness of the matches in ranking
instantiations, and captures the response generation process in an
overall collaborative framework that can negotiate proposals with the
user.

\section{Conclusions and Future Work}

This paper has presented a plan-based system that captures
collaborative response generation in a {\em Propose-Evaluate-Modify}
cycle. Our system can initiate subdialogues to negotiate implicitly
proposed additions to the shared plan, can appropriately respond to
user queries that are motivated by ill-formed or suboptimal solutions,
and handles in a unified manner the negotiation of proposed domain
actions, proposed
problem-solving actions, and beliefs proposed by discourse actions.
In addition, our system captures cooperative
responses within an overall collaborative framework that allows for
negotiation and accounts for why questions are sometimes never
answered (even in the most cooperative of environments).

This response generation architecture has been implemented in a
prototype system for a university advisement domain. The system is
presented with the existing dialogue model and the actions proposed by
the user's new utterances. It then produces as output the logical form
for the appropriate collaborative system response. In the future, we
will extend our system to include various argumentation strategies
\cite{syc_ijcai89,qui_phd,may_aaai93} for supporting its claims.

\section{Acknowledgments}

The authors would like to thank Stephanie Elzer for her comments on
earlier drafts of this paper.

\nocite{sid_aaaiws92}

\end{document}